%% file: main.tex
\documentclass[10pt,twocolumn,letterpaper]{article}

\usepackage{cvpr}              
\usepackage{cuted}
\usepackage{capt-of}
\usepackage{todonotes}
\usepackage{arydshln}


\definecolor{cvprblue}{rgb}{0.21,0.49,0.74}
\usepackage[pagebackref,breaklinks,colorlinks,citecolor=cvprblue]{hyperref}

\def\paperID{52} 
\def\confName{CVPR}
\def\confYear{2024}

\title{Gaussian Splatting Decoder for 3D-aware Generative Adversarial Networks}

\author{
Florian  Barthel\textsuperscript{1, 2}
\quad Arian Beckmann\textsuperscript{1}
\quad Wieland Morgenstern\textsuperscript{1}
\quad Anna Hilsmann\textsuperscript{1} 
\quad Peter Eisert\textsuperscript{1,2} 
\\[0.2cm]
\textsuperscript{1} Fraunhofer Heinrich Hertz Institute, HHI \\ \textsuperscript{2} Humboldt University of Berlin
\\[0.2cm]
}

\begin{document}

\maketitle

\begin{strip}
    \centering
    \includegraphics[width=0.86\textwidth]{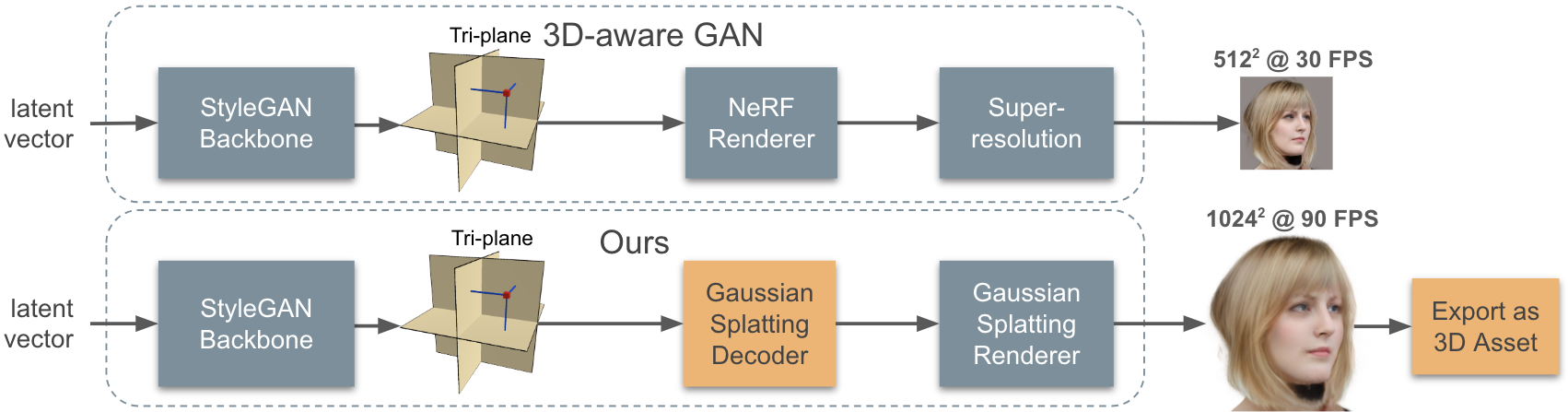}
    \vspace{-0.2cm}
    \captionof{figure}{We propose a novel 3D Gaussian Splatting decoder that converts high quality results from pre-trained 3D-aware GANs into Gaussian Splatting scenes in real-time for efficient and high resolution rendering.
    }
    \label{fig:feature-graphic}
\end{strip}

\input{sec/0_abstract}    
\input{sec/1_intro}
\input{sec/2_related_work}

\input{sec/3_method}

\input{sec/4_experiments}
\input{sec/5_conclusion}

{
    \small
    \bibliographystyle{ieeenat_fullname}
    \bibliography{main}
}

\input{sec/X_suppl}

\end{document}

%% file: sec/0_abstract.tex
\begin{abstract}
\vspace{-0.5cm}
NeRF-based 3D-aware Generative Adversarial Networks (GANs) like EG3D or GIRAFFE have shown very high rendering quality under large representational variety. However, rendering with Neural Radiance Fields poses challenges for 3D applications: First, the significant computational demands of NeRF rendering preclude its use on low-power devices, such as mobiles and VR/AR headsets. Second, implicit representations based on neural networks are difficult to incorporate into explicit 3D scenes, such as VR environments or video games. 3D Gaussian Splatting (3DGS) overcomes these limitations by providing an explicit 3D representation that can be rendered efficiently at high frame rates. In this work, we present a novel approach that combines the high rendering quality of NeRF-based 3D-aware GANs with the flexibility and computational advantages of 3DGS. By training a decoder that maps implicit NeRF representations to explicit 3D Gaussian Splatting attributes, we can integrate the representational diversity and quality of 3D GANs into the ecosystem of 3D Gaussian Splatting for the first time. Additionally, our approach allows for a high resolution GAN inversion and real-time GAN editing with 3D Gaussian Splatting scenes.
\\
Project page: \href{https://florian-barthel.github.io/gaussian_decoder/}{florian-barthel.github.io/gaussian\_decoder}

\end{abstract}

%% file: sec/1_intro.tex
\section{Introduction}
\label{sec:intro}

Creating and editing realistic 3D assets is of vital importance for applications such as Virtual Reality (VR) or video games. Often, this process is very costly and requires a significant amount of manual editing. Over the last few years, there have been drastic improvements to 2D \cite{karras_progressive_2018,karras_style-based_2019,karras_analyzing_2020,stylegan_xl} and 3D \cite{chan_efficient_2022,niemeyer_giraffe_2021,chan_pi-gan_2021,Brehm2022,shi_lifting_2021,an_panohead_2023,Xu_omniavatar_2023_CVPR} image synthesis. 
These advancements increasingly narrow the gap between professionally created 3D assets and those that are automatically synthesized. One of the most notable recent methods is the \textit{Efficient Geometry-aware 3D GAN (EG3D)}  \cite{chan_efficient_2022}. It successfully combines the strength of \textit{\mbox{StyleGAN}} \cite{karras_analyzing_2020}, originally built for 2D image generation, with a 3D NeRF renderer \cite{mildenhall_nerf_2020,barron_mip-nerf_2021}, achieving state-of-the-art 3D renderings synthesized from a latent space. 
Despite EG3D's significant contributions to 3D rendering quality, its integration into 3D modeling environments like Unity or Blender remains difficult. This challenge stems from its NeRF dependency, which only generates 2D images from 3D scenes, without ever explicitly representing the 3D scene. As a result, EG3D cannot be imported or manipulated in these computer graphics tools.



Recently introduced, \textit{3D Gaussian Splatting (3DGS)} \cite{kerbl3Dgaussians} provides a novel explicit 3D scene representation, enabling high-quality renderings at high frame rates.
Following its debut, numerous derivative techniques have already emerged \cite{chen2024survey}, including the synthesis of controllable human heads \cite{qian2023gaussianavatars,xu2023gaussianheadavatar,Zhao2024psavatar}, the rendering of full body humans \cite{hugs} or the compression of the storage size of Gaussian objects \cite{morgenstern2023compact}. 
On the one hand, 3DGS provides a substantial improvement in terms of rendering speed and flexibility compared to NeRF: The explicit modelling enables simple exporting of the scenes into 3D software environments. Furthermore, the novel and efficient renderer in 3DGS allows for high-resolutions renderings, and an increase in rendering speed with a factor of up to $1000 \times$ over state-of-the-art NeRF frameworks \cite{3dgs, Barron2021MipNeRF3U, muellerinstant, yu_and_fridovichkeil2021plenoxels}. 
On the other hand, NeRF's implicit scene representation allows for straightforward decoding of scene information from latent spaces. Notably through the usage of tri-planes \cite{chan_efficient_2022}, which store visual and geometric information of the scene to be rendered. This enables the integration of NeRF rendering into GAN frameworks, lifting the representational variety and visual fidelity of GANs up into three-dimensional space.  Combining NeRFs and GANs is highly advantageous, as rendering from a latent space offers multiple benefits: Firstly, it allows for rendering an unlimited amount of unique appearances. Secondly, a large variety of editing methods \cite{harkonen_ganspace_2020,styleflow,mirza_conditional_2014} can be applied. And thirdly, single 2D images can be inverted, using 3D GAN inversion \cite{karras_style-based_2019,roich_pivotal_2021,Barthel2024}, allowing for full 3D reconstructions from a single image.

Sampling visual information from latent spaces with large representational variety poses a challenge for rendering with 3DGS, as the framework requires the information for the appearance of the scene to be encoded as attributes of individual splats, rather than in the latent space itself. This severely complicates the task of fitting 3D Gaussian splats to variable latent spaces, given that the splats would need to be repositioned for every new latent code - a challenge that is not addressed in the original 3DGS framework. Several approaches tackling the problem of rendering with 3DGS from latent tri-planes have been proposed \cite{hugs,triplane_meets_gaussian,lan2023gaussian3diff}, but to the best of our knowledge, no method exists to create 3D heads rendered with Gaussian Splatting from a latent space.

In this work, we propose a framework for the synthesis of explicit 3D scenes representing human heads from a latent space. This is done by combining the representational variety and fidelity of 3D-aware GANs with the explicit scenes and fast rendering speed of 3D Gaussian Splatting. Our main contributions can be summarized as follows:


\begin{enumerate}

    \item A novel method that allows for GAN-based synthesis of explicit 3D Gaussian Splatting scenes, additionally avoiding superresolution modules as used in the generation of implicit scene representations.
    \item A novel sequential decoder architecture, a strategy for sampling Gaussian splat positions around human heads and a generator backbone fine-tuning technique to improve the decoders capacity.
    \item An open source end-to-end pipeline for synthesizing state-of-the-art 3D assets to be used in 3D software.

\end{enumerate}

%% file: sec/2_related_work.tex
\section{Related Work}
\label{sec:related_work}

\subsection{Neural Radiance Fields}

In their foundational work on Neural Radiance Fields (NeRFs) Mildenhall \etal \cite{nerf} propose to implicitly represent a 3D scene with an MLP that outputs color and density at any point in space from any viewing direction. This representation revolutionized novel-view synthesis, due to its ability to reconstruct scenes with high fidelity, high flexibility with respect to viewpoints, and compactness in representation through the usage of the MLP. To obtain a 2D rendering, a ray is cast for each pixel from the camera into the scene with multiple points sampled along each ray, which in turn are fed to the MLP in order to obtain their respective color and density values. 
NeRFs have proven to provide high-quality renderings, but are slow during both training and inference: the sampling and decoding process require querying a substantial number of points per ray, which has a high computational cost.
Successors of the seminal NeRF approach are subject to improvements in quality \cite{martinbrualla2020nerfw} as well as training and inference speed \cite{barron_mip-nerf_2021, yu_and_fridovichkeil2021plenoxels, muellerinstant}. Plenoxels \cite{yu_and_fridovichkeil2021plenoxels} replaces the MLP representation of the scene by a sparse voxel grid representation, which leads to a speed-up in optimization time by two orders of magnitude compared to vanilla NeRF while maintaining high rendering quality. InstantNGP \cite{muellerinstant} proposes the usage of multi-resolution hash tables to store scene-related information. 
Through leveraging parallelism on GPUs and a very optimized implementation that fits the hash tables into the GPU cache, it achieves significant speedups in processing times, making real-time applications feasible.

Moreover, various approaches aiming to render in real-time propose to either store the view-dependent colors and opacities of NeRF in volumetric data representations or partition the scene into multiple voxels represented by small independent neural networks \cite{zhangdigging, yu2021plenoctrees, lee2023mfnerf, fastnerf, Sun2021DirectVG, tensorrf, relufields, NEURIPS2020_b4b75896}.

\subsection{3D Gaussian Splatting}
Recently, Kerbl \etal \cite{3dgs} proposed to represent scenes explicitly in the form of Gaussian splats. Each singular splat represents a three-dimensional Gaussian distribution with mean $\mu$ and covariance matrix $\Sigma$. For computational simplicity, the authors decide to represent the covariance matrix as the configuration of an ellipsoid, i.e.\ $\Sigma=RSS^{T}R^{T}$, with scaling and rotation matrices $S$ and $R$. To characterize the appearance, each splat holds further attributes describing its opacity and view-dependent color through a set of spherical harmonics. Each splat's attributes are optimized in end-to-end training, utilizing a novel differentiable renderer. This renderer is essential for the success of 3D Gaussian Splatting. Its architectural design allows for high-resolution rendering in real-time through fast GPU execution utilizing anisotropic splatting while being visibility-aware. 
This architectural design significantly accelerates the training process and novel-view rendering time. In general, 3DGS \cite{3dgs} is able to outperform several NeRF-based state-of-the-art approaches in rendering speed by factors of up to $1000 \times$, while keeping competitive or better image quality.


Several approaches that utilize 3DGS for the representation and rendering of human heads have been proposed \cite{Zhao2024psavatar, qian2023gaussianavatars, xu2023gaussianheadavatar}. GaussianAvatar \cite{qian2023gaussianavatars} allows editing of facial expressions within a scene of Gaussians already fitted to a specific identity. To do so, they use the FLAME \cite{FLAME:SiggraphAsia2017} 3D morphable face model to create a triangle-mesh representing the head in 3D space and assign a splat to each triangle. Moreover, the geometric attributes of the splats are dynamically adjusted to align with the respective triangle's properties; for example, the global rotation of the splat is adjusted to match that of the triangle. 
Similarly, Xu \etal \cite{xu2023gaussianheadavatar}
anchor the Gaussian splats to a 3D triangle mesh fitted to a head that depicts a neutral expression. They utilize deformation MLPs conditioned on expression vectors to adjust the triangle-mesh and the resulting Gaussian positions to account for changes in expression. Ultimately, they render a feature map from their scene with 3DGS and translate those into high-fidelity head images in 2K resolution with a super-resolution network.

\begin{figure*}[htbp]
    \centering
    \includegraphics[width=0.95\linewidth]{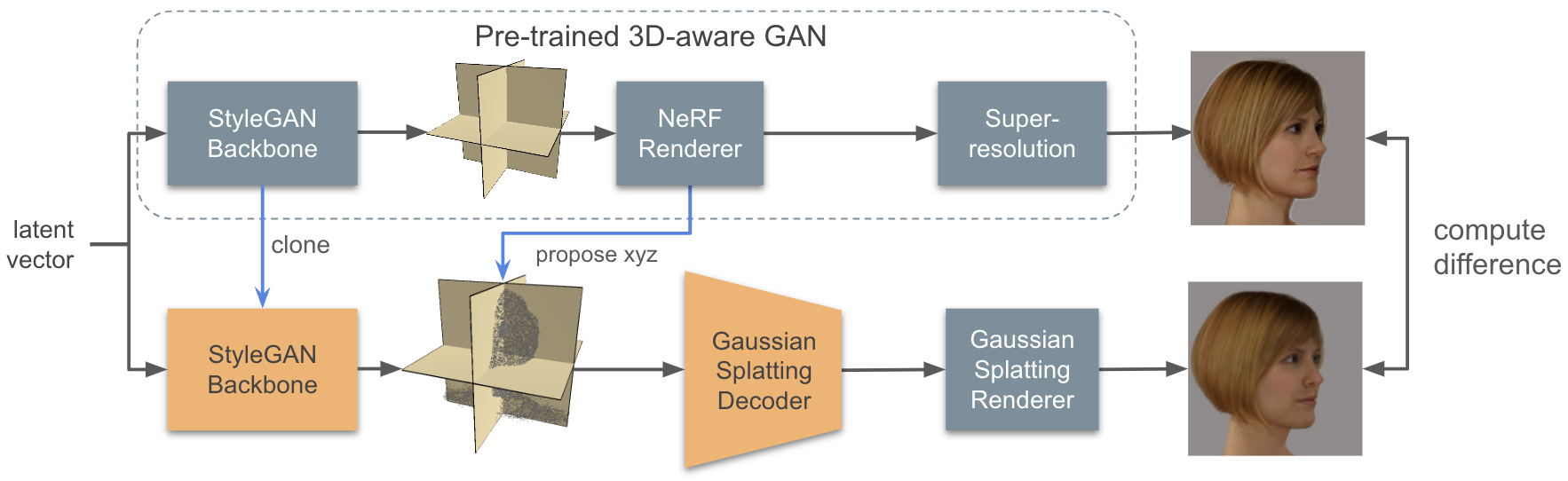}
    \caption{Visualization of our method (orange parts are optimized). We initially clone the backbone of the 3D-aware GAN. Afterwards, we iteratively optimize the Gaussian Splatting decoder by comparing the output of the pre-trained GAN, after super-resolution, with the output of the decoder. The xyz coordinates at which the tri-plane is sampled originates from the density information of the NeRF renderer.}
    \label{fig:method_overview}
\end{figure*}

\subsection{3D-aware GANs}
Following the success of 2D GANs in recent years, several methods have been proposed to synthesize 3D content with GANs as well. To achieve this, the generator component of a GAN is modified to create an internal 3D representation suitable for output through differentiable renderers. Given that these renderers return a 2D representation of the 3D model, the framework can be trained with 2D data. This is crucial, as high-quality 2D datasets are more readily available compared to their 3D counterparts. One of the first high-resolution 3D-aware GANs is \textit{LiftingStyleGAN} \cite{Shi2020Lifting2S}. Its architecture extends the 2D \textit{StyleGAN} \cite{karras_alias-free_2021} with a custom built renderer based on texture and depth maps. Shortly after, $\pi$-\textit{GAN} \cite{chan_pi-gan_2021} and \textit{GIRAFFE} \cite{niemeyer_giraffe_2021} have been introduced that both use a Neural Radiance Field (NeRF) renderer. Those methods show very promising visual results. Nevertheless, while $\pi$-GAN is very slow in rendering, only achieving 1 frame per second, GIRAFFE fails to estimate a good 3D geometry. Both challenges are solved with the introduction of the \textit{Efficient Geometry-aware 3D Generative Adversarial Network (EG3D)} by Chan \etal \cite{chan_efficient_2022}. EG3D combines the strength of the StyleGAN architecture for 2D image synthesis with the rendering quality of a NeRF renderer. This is done by reshaping the output features of a StyleGAN generator into a three-dimensional tri-plane structure to span a 3D space. From this tri-plane, 3D points are projected onto 2D feature maps and forwarded to a NeRF renderer. The renderer creates a 2D image at a small resolution, which is then forwarded to a super-resolution module. This approach returns state-of-the-art renderings at a resolution of 512x512 pixels, while maintaining reasonable rendering speeds of about 30 FPS. The use of the super-resolution module effectively locks in the output size and aspect ratio, making it impossible to adjust or enlarge them without training a completely new network. This limitation of EG3D contrasts with explicit methods, which feature a renderer capable of adjusting the rendering resolution on demand.

Since the inception of EG3D, several approaches have adopted its architecture, extending the rendering capabilities \cite{Xu_omniavatar_2023_CVPR,li2023instant3d,an_panohead_2023,nvdiffrec}. 
\textit{PanoHead} \cite{an_panohead_2023}, stands out in particular, as it addresses synthesizing full 360° heads. This is done by adding further training data that shows heads from the back and by disentangling the head from the background.  The latter is done by separately generating the background and blending it with the foreground using a foreground mask obtained during the rendering process.

\subsection{Decoding Gaussian Attributes from Tri-planes}

Decoding NeRF attributes, i.e.\ color and density, from a tri-plane has proven to produce state-of-the-art frameworks. Decoding tri-planes into Gaussian Splatting attributes, on the other hand, induces further complexity. This is because a Gaussian splat, located at a specific position on the tri-plane need not represent the color and density of this specific location, but instead a 3D shape with a scale that extends into other regions of the scene. Naively, this could be solved by treating Gaussian splats as a point cloud with a very high number of tiny colored points. This approach would however neglect the advantages of Gaussian Splatting and reintroduce high computational costs for rendering the point cloud. Instead, when decoding Gaussian attributes \cite{3dgs}, we seek to find suitable representations, such that Gaussian splats adapt their geometry to represent the structure of the target shape. Thus, smooth surfaces should be represented by wide flat Gaussian splats, while fine structures are best represented by thin long Gaussians.

Recent work already investigated the ability to decode Gaussian splats from tri-planes. \textit{HUGS} \cite{hugs} uses a small fully connected network to predict the Gaussian attributes to render full body humans in 3D. Contrary to our approach, HUGS overfits a single identity iteratively instead of converting any IDs from a latent space in a single shot. Similarly, \cite{triplane_meets_gaussian} uses a transformer network to decode Gaussian attributes from a tri-plane in order to synthesize 3D objects. A different approach that also combines 3DGS with tri-planes is \textit{Gaussian3Diff} \cite{lan2023gaussian3diff}. Instead of decoding Gaussian attributes from a tri-plane, they equip each Gaussian with a local tri-plane that is attached to its position. This hybrid approach shows promising quality, although the rendering speed is lower compared to 3DGS.




%% file: sec/3_method.tex
\section{Method}
\label{sec:method}

Our goal is to design a decoder network that converts the output of a 3D-aware GAN, specifically tailored for human head generation, into a 3D Gaussian Splatting scene without requiring an iterative scene optimization process. An overview of our method is shown in Figure \ref{fig:method_overview}. We extract the tri-plane of the 3D GAN, which is originally used to render a NeRF scene, and train a decoder network to obtain 3D Gaussian Splatting attributes (i.e.~position, color, rotation, scale, and opacity). For simplicity, we omit the estimation of view-dependent spherical harmonics coefficients. For training, we compare the synthesized images from the 3D GAN to the renderings of the decoded 3D Gaussian Splatting scenes. Importantly, our decoder does not use any super-resolution module. Instead, we render the decoded Gaussian Splatting scene already at the same resolution as the final output of the 3D GAN. The absence of a super-resolution module allows the export of the decoded scene directly into 3D modeling environments, and for rendering at different resolutions and aspect ratios at high frame rates. 

\subsection{Position Initialization}
\label{subsec:pos-init}
Given that the 3D Gaussian splats, being described with multiple attributes (position, color, rotation, scale, and opacity), have multiple degrees of freedom, it is difficult to receive a meaningful gradient for the position during optimization. To overcome this issue, 3DGS uses several strategies to prune, clone, and split Gaussians during the training in order to spawn new Gaussians at fitting locations in the scene or remove redundant ones. For example, if a Gaussian splat is located at an incorrect position, 3DGS prefers to make the Gaussian splat vanish by reducing the opacity or to change its color to fit the current position, rather than moving its position. For our purpose of training a decoder that automatically creates new Gaussian scenes in a single forward pass, this iterative approach is not available. Instead, we take advantage of the geometric information already contained in the pre-trained 3D GAN's tri-plane. This is done by decoding the tri-plane features into opacity values using the pre-trained MLP of the NeRF renderer followed by a surface estimation based on the opacity values. Specifically, we uniformly sample a cube of points ($128 \times 128 \times 128$), decode the opacity and estimate the surface with marching cubes \cite{marchingcubes}. On this surface, we sample 500k points at random positions and slightly interpolate the points randomly towards the center, thus creating a thick surface.
This provides us with a good position initialization for the Gaussians representing any head created by the 3D GAN. Even so, sampling the opacity from the NeRF renderer is computationally expensive. Nevertheless, this only has to be done once per ID / latent vector. After the 3D Gaussian scene is created, it can be rendered very efficiently.

\subsection{Decoder Architecture}
\label{subsec:dec-arch}
Recent work that use a decoder network to estimate Gaussian Splatting attributes from tri-plane features use fully connected networks \cite{hugs} or transformer-based models \cite{triplane_meets_gaussian}. For our approach, we also use a fully connected network, however, instead of computing all Gaussian attributes at once, we calculate them sequentially. Specifically, we first forward the tri-plane features to the first decoder that estimates the color. After that, we use the information of the color together with the tri-plane features and feed them to the next decoder that estimates the opacity. This is done iteratively until all attributes are decoded (color $\rightarrow$ opacity $\rightarrow$ rotation $\rightarrow$ scale $\rightarrow$ position offset). Thus, the last decoder receives all preceding attributes along with the tri-plane features. The intuition behind this approach is to create a dependency between the attributes. We hypothesize that, for instance, the scale decoder benefits from information about the color or rotation, in order to decide how large the respective Gaussian splat will be. Additionally, the high degrees of freedom of the combined Gaussian splat attributes get reduced heavily for each decoder, allowing for easier specialization.

Inside each decoder, we use three hidden layers, each equipped with 128 neurons and a \textit{GELU} activation. The output layer has no activation function, except for the scaling decoder. There, we apply an inverted \textit{Softplus} activation to keep the splats from getting too large, avoiding excessive GPU memory usage during rasterization.

\begin{figure}[htbp]
    \centering
    \includegraphics[width=\linewidth]{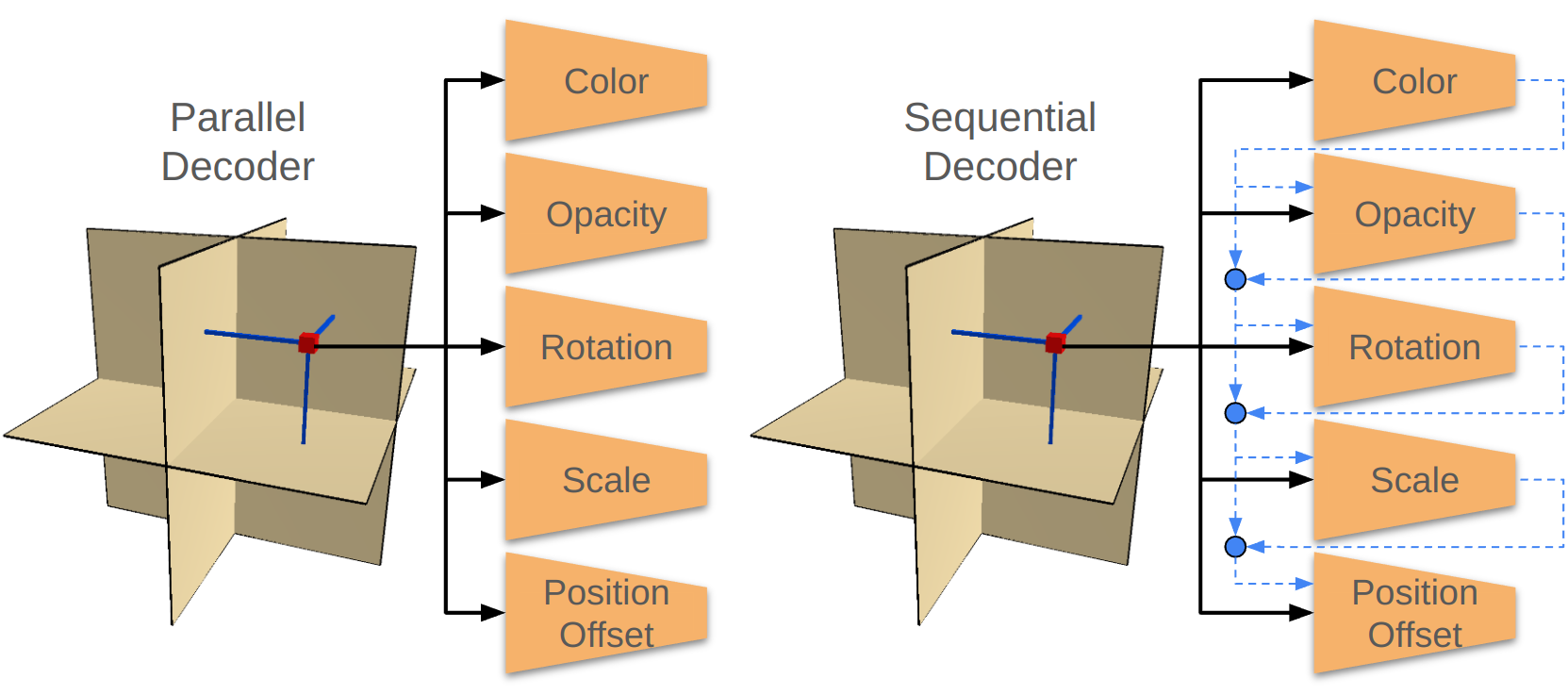}
    \caption{A comparison between a parallel decoder that maps all Gaussian attributes at once to our sequential decoder, where each attribute is decoded after another using the prior information. }
    \label{fig:decoder_arch}
\end{figure}

\begin{table*}
    \centering
    \begin{tabular}{|l|l|c|c|c|c|cc|}
        \hline
        Model & Training Data & MSE $\downarrow$ & LPIPS $\downarrow$ & SSIM $\uparrow$ & ID Sim. $\uparrow$ & FPS@512 $\uparrow$ & FPS@1024 $\uparrow$ \\
        \hline
        
        PanoHead & FFHQ-H &  & & & & 37 & N/A\\
        Ours & PanoHead & 0.002 & 0.161 & 0.820 & 0.902 & 170 & 90 \\
        \hline
        
        $\text{EG3D}_{\text{LPFF}}$ & LPFF &  & & & & 32 & N/A \\
        Ours & $\text{EG3D}_{\text{LPFF}}$ & 0.004 & 0.248 & 0.852 & 0.946 & 135 & 96 \\
        \hline
        
        $\text{EG3D}_{\text{FFHQ}}$ & FFHQ & & & & & 31 & N/A \\
        Ours & $\text{EG3D}_{\text{FFHQ}}$ & 0.002 & 0.195 & 0.842 & 0.968 & 164 & 132 \\
        \hline
    \end{tabular}
    \caption{Results for training our decoder for different pre-trained 3D-aware GANs. Columns with \textit{Ours} refer to a decoder that was trained with the GAN specified in the column above. For all three decoders we observe high similarity scores along with high rendering speeds.
    }
    \label{tab:general_metrics}
\end{table*}

\subsection{Backbone Fine-tuning}
\label{subsec:backbone}
In addition to optimizing the weights of the decoder network, we create a copy of the pre-trained 3D generator and optimize its weights as well. This fine-tuning allows the optimization process to adapt the tri-plane features to provide a better basis for creating Gaussian Splatting attributes, as they are inherently different. While NeRFs only require the color and density of a specific location, Gaussian splats additionally have a scale and rotation, thus influencing adjacent regions too.


\subsection{Loss Functions}
\label{subsec:losses}
The vanilla 3D Gaussian Splatting algorithm uses a combination of L1 loss and structural similarity. This combination has proven to be very successful for learning static scenes. For our purpose, however, of learning a decoder network that is able to synthesize an tremendous diversity of images, it requires a loss function that provides better perceptual feedback. This is because we aim to produce a 3D Gaussian Splatting face that looks perceptually very close to the GAN rendered face, without penalizing the model too much if small structures like hair do not align perfectly. For that reason, we supplement the existing L1 and structural similarity loss with an LPIPS norm \cite{lpips} and an ID similarity \cite{arcface} loss. This ID loss is based on a pre-trained face detector (ArcFace) and estimates how similar two faces are. Since PanoHead renders the head from all 360° views, we only apply the ID loss, when the face is viewed from a frontal viewpoint. Additionally, to guide the decoder towards areas needing finer structural details, we calculate the difference between the synthesized image and target image after applying a Sobel filter. Formally, our loss function can be expressed as follows:

\begin{equation}
    \mathcal{L} = \lambda_{1}\mathcal{L}_\text{L1} + \lambda_{2}\mathcal{L}_\text{SSIM} + \lambda_{3}\mathcal{L}_\text{LPIPS} + \lambda_{4}\mathcal{L}_\text{ID} + \lambda_{5}\mathcal{L}_\text{Sobel}.
\end{equation}




%% file: sec/4_experiments.tex
\section{Experiments}
\label{sec:experiments}

\subsection{Implementation Details}
In the following experiments, we train our Gaussian splatting decoder for multiple pre-trained target GANs. These are: EG3D trained on the FFHQ \cite{karras_style-based_2019} dataset, EG3D trained on the LPFF dataset \cite{wu_lpff_2023}, and PanoHead trained on the FFHQ-H \cite{an_panohead_2023} dataset. We train for 500k iterations with an Adam optimizer using a learning rate of 0.0009. Loss weights are set to $(\lambda_{1}, \lambda_{2}, \lambda_{3}, \lambda_{4}, \lambda_{5}) = (0.2, 0.5, 1.0, 1.0, 0.2)$ for all experiments unless stated otherwise. For PanoHead, we sample random cameras all around the head, and for EG3D, we sample mainly frontal views with small vertical and horizontal rotations.

We have optimized all training parameters for PanoHead, since it synthesizes full 360° views, making it ideally suited for being rendered in an explicit 3D space.

\subsection{Metrics}
To evaluate the performance of our decoder, we measure the image similarity for 10k images using MSE, LPIPS, ID similarity and structural similarity. 
In order to measure the frame rate, we use a custom visualization tool that is based on the EG3D visualizer. This way, we ensure that the performance differences are purely due to the renderer and not dependent on the programming language or compiler. With very efficient Gaussian splatting renderer like the SIBR viewer \cite{kerbl3Dgaussians} that is purely written in C++, we could achieve even higher FPS.

\subsection{Quantitative Results}

After training our decoders, we observe a high image similarity between the decoded Gaussian Splatting scene and the respective target GAN as stated in Table \ref{tab:general_metrics}. The low MSE and SSIM indicate that the renderings have similar colors and structures, respectively. In addition, the LPIPS and ID similarity metrics underline that the images are perceptually very close. The highest ID similarity is found when decoding the $\text{EG3D}_{\text{FFHQ}}$ model. Here, we reach a similarity score of 0.968. A possible explanation for this is that the FFHQ training dataset contains the fewest images across all three comparisons, making it the easiest to decode, given that there is less variation. The lowest ID similarity is found for the decoder trained with the PanoHead GAN. In this case, the decoder has to learn a full 360° view of the head. This is not regarded by the ID similarity as it is only computed for renderings showing a frontal view.

Considering render speed for each model, rendering the Gaussian Splatting scene achieves about four times the FPS compared to rendering the 3D-aware GANs. Furthermore, as the rendering resolution for Gaussian Splatting is not limited by any super-resolution module, it can be rendered at arbitrary resolution. Here, we observe that when increasing resolution four-fold, we still achieve more than three times the framerate of the GAN models at the lower resolution.

\begin{figure}[b]
    \centering
    \includegraphics[width=0.9\linewidth]{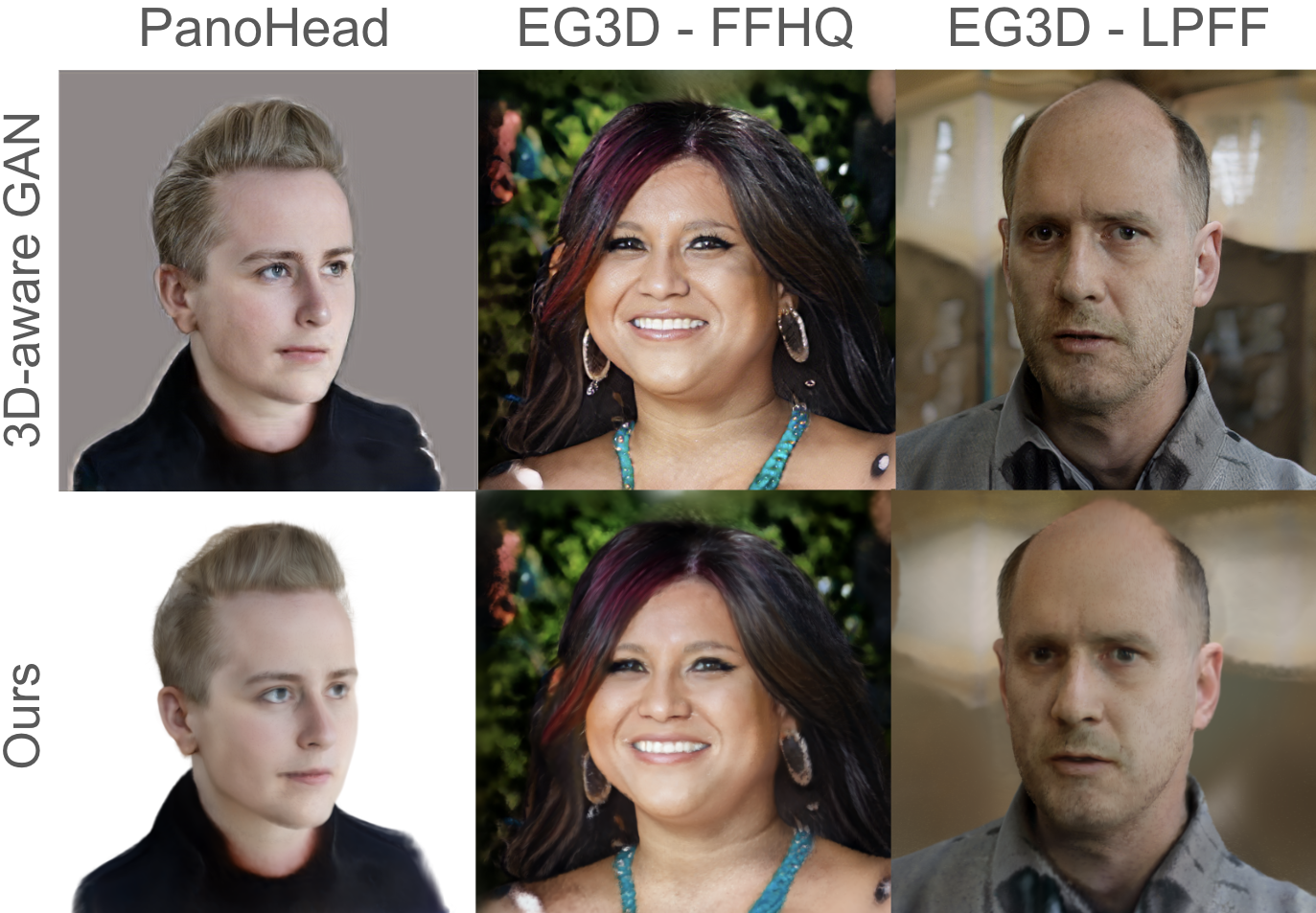}
    \caption{Example renderings of the target images produced by respective 3D-aware GAN (top row) and the renderings of the decoded 3D Gaussian Splatting scene (bottom row, \textit{Ours}). Additional renderings can be found in the supplementary material.}
    \label{fig:example_renderings}
\end{figure}

\begin{figure}[htbp]
    \centering
    \includegraphics[width=1\linewidth]{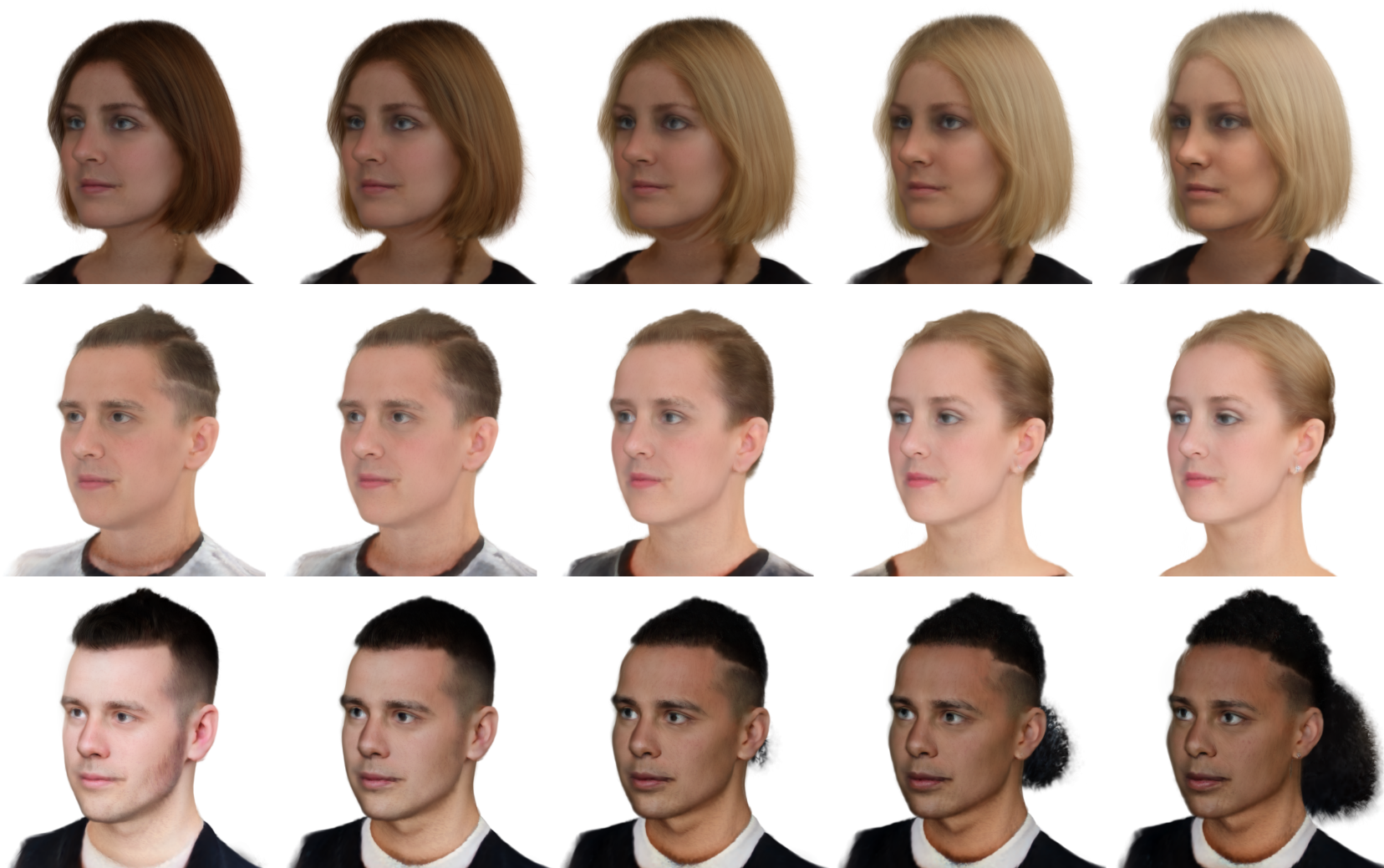}
    \caption{Renderings for example interpolating paths, demonstrating the possibility for applying GAN editing methods.}
    \label{fig:interpolate}
\end{figure}

\begin{figure*}
    \centering
    \includegraphics[width=0.9\linewidth]{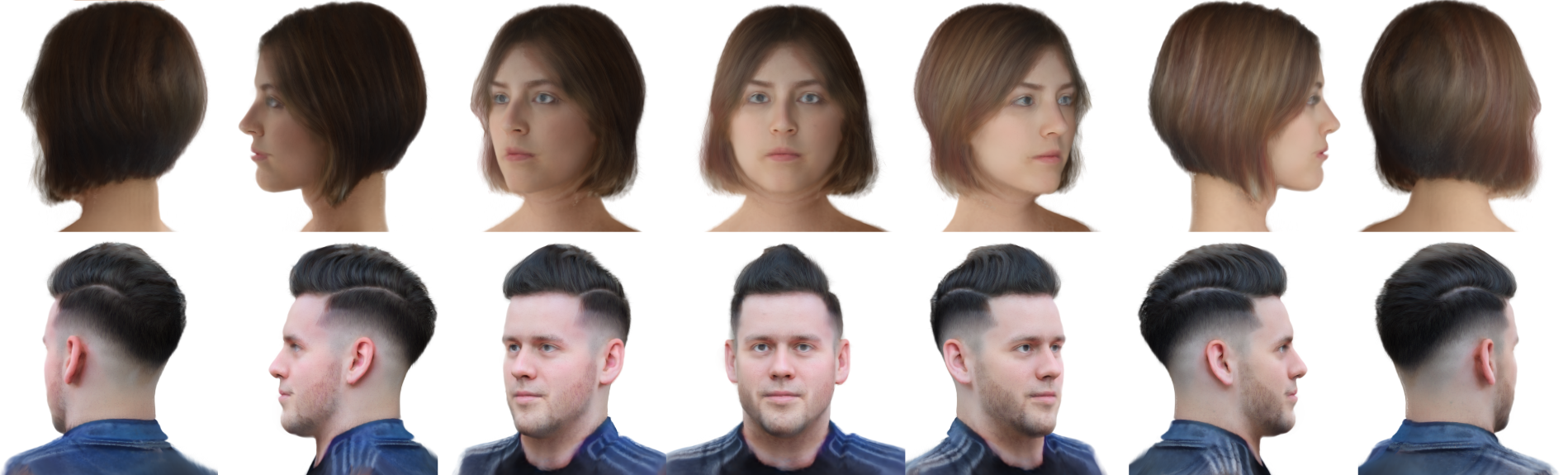}
    \vspace{-0.1cm}
    \caption{Example 360° rendering of our Gaussian Splatting decoder trained with PanoHead.}
    \label{fig:example_panohead}
\end{figure*}

\subsection{Qualitative Results}
\label{sec:qaulitative}
In addition to the quantitative measures, we also demonstrate our method qualitatively. Figure \ref{fig:example_renderings} shows one example rendering for each of the three 3D-GAN methods, along with our decoded Gaussian Splatting scenes. We observe a high visual similarity between target and rendering as indicated by the image similarity metrics. While EG3D uses one single 3D scene that combines head and background, PanoHead uses two separate renderings. This allows our decoder to exclusively learn the head and rotate it independently to the background. An example of a full 360° rotation of two decoded Gaussian Splatting heads is shown in Figure \ref{fig:example_panohead}.

Additionally, we observe a reduction in aliasing and texture sticking artifacts with our 3D representation when rotating the camera around the head. This was often observed when rendering with EG3D or PanoHead. Specifically, some structures like hair or skin pores shifted when chaining the camera viewpoint, instead of moving along with the 3D head. This is no longer the case with our Gaussian Splatting representation, as we produce one fixed explicit 3D scene for each ID. 

We also demonstrate in Figure \ref{fig:interpolate} that our decoder allows latent interpolation. This opens up various GAN editing or GAN inversion methods to be applicable to our method.

In some renderings, we observe that the eyes appear uncanny or blurry. We believe this occurs because the underlying target data produced from the 3D-aware GAN almost exclusively shows renderings where the person is looking towards the camera. 
The GAN's NeRF-renderer, being view-dependent, likely learns to place the pupils to align with the camera angle. However, as we disabled the spherical harmonics to reduce complexity, our decoder is not able to learn any view dependencies. Instead, it learns an averaged eye, which is slightly blurry and always looks forward. To overcome this limitation, it might be beneficial to incorporate the spherical harmonics into the decoder training for future work.

\subsection{Ablation Study}
\label{sec:ablation}
In the following, we will justify our design decisions by performing an ablation study to the main components. 

\begin{figure}[ht]
    \centering
    \includegraphics[width=0.9\linewidth]{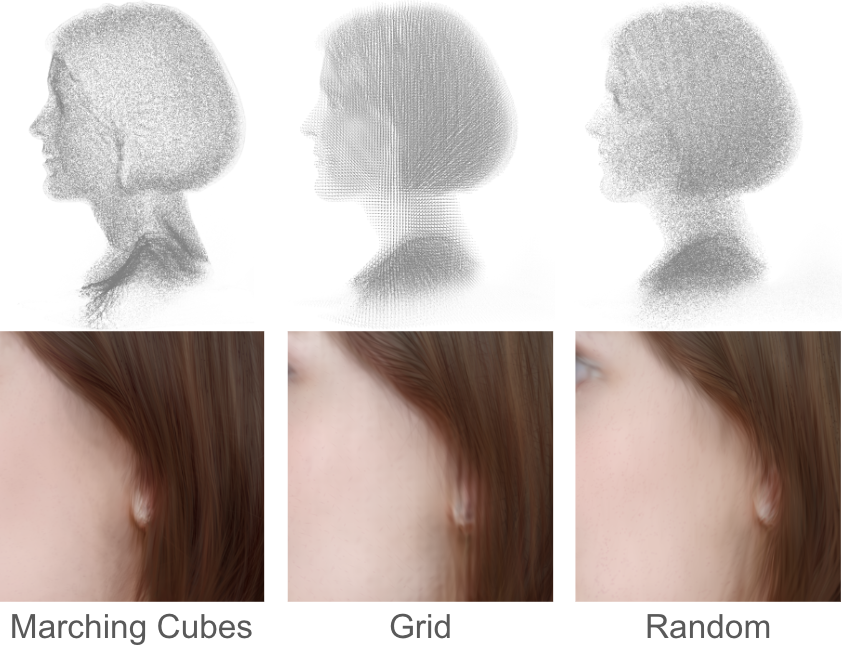}
    \vspace{-0.1cm}
    \caption{A visual comparison between different position sampling approaches. When using grid sampling, we observe some grid artifacts on the surface.}
    \label{fig:position_init_compare}
\end{figure}

\noindent{\textbf{Position Initialization:}} The position initialization is a crucial component of our decoder as it decides where to place the Gaussian splats. For this, multiple approaches are possible: Sampling the points on a 3D grid, sampling the points randomly in a 3D cube or sampling the points on the surface of a 3D shape, created by marching cubes. Interestingly, when looking at the quantitative results in for all three approaches in Table \ref{tab:pos_sampling}, we clearly favor sampling on a 3D grid, as it achieves the overall best scores. Nevertheless, when inspecting the resulting renderings in Figure \ref{fig:position_init_compare}, we observe that grid sampling creates some artifacts. We see some horizontal and vertical lines on the surface of the head, where the splats are placed. Although this is not penalized by the chosen metrics, it significantly decreases the level of realism. Therefore, given that the marching cube sampling scores second best, while producing good visual results, we chose it for our decoder.

\begin{table}[h]
    \centering
    \begin{tabular}{|l|ccc|}
        \hline
        Sampling Method & LPIPS $\downarrow$ & SSIM $\uparrow$ & ID Sim $\uparrow$ \\
        \hline
        Random Pos          &	0.179	&	0.839	&	0.856 \\
        3D Grid          &	\textbf{0.167}	&	\textbf{0.851}	&	\textbf{0.898} \\
        March. Cubes    &	0.176	&	0.842	&	0.883\\
        \hline
    \end{tabular}
    \caption{Comparing different position sampling methods. 
    }
    \label{tab:pos_sampling}
\end{table}

\noindent{\textbf{Decoder Architecture:}} The core component of our method is the decoder. Its architecture can have a big influence on the capacity to learn a mapping between tri-plane features and Gaussian Splatting attributes. In the following, we will look into three different architecture types. First, the sequential decoder, decoding each attribute with the information of the previous one (color $\rightarrow$ opacity $\rightarrow$ rotation $\rightarrow$ scale $\rightarrow$ position offset), a parallel decoder that maps all attributes at once and again a sequential decoder where we invert the order of the decoders. The results for all three decoder are listed in Table \ref{tab:architectures}. We notice that the sequential decoder is the overall best, although with a very slim margin to the parallel decoder. Interestingly, despite having the same amount of connections in the network as the parallel decoder, the reversed sequential decoder performs worse, suggesting that the order of decoding significantly impacts performance. A possible explanation for this disparity is that the outputs from earlier stages in the sequential version might introduce noise, thereby impeding the optimization process.

\begin{table}[htbp]
    \centering
    \begin{tabular}{|l|ccc|}
        \hline
        Architecture & LPIPS $\downarrow$ & SSIM $\uparrow$ & ID Sim $\uparrow$ \\
        \hline
        Sequential          &	\textbf{0.176}	&	\textbf{0.842}	&	\textbf{0.883} \\
        Parallel            &	0.177	&	0.841	&	0.879\\
        Sequential Reversed &	0.228	&	0.803	&	0.765\\
        \hline
    \end{tabular}
    \caption{Image difference metrics for decoder architectures.}
    \label{tab:architectures}
\end{table}

\noindent{\textbf{Backbone Fine-tuning:}} During the training, we fine-tune the weights of the pre-trained StyleGAN backbone that produces the tri-plane features. This distributes some of the work load from the decoder to the backbone, as we penalize the tri-plane creation if the decoder cannot easily create Gaussian splats from it. Disabling this component leads to a decline in all performance metrics, especially the ID similarity, which drops from 0.883 to 0.858 as shown in Table~\ref{tab:finetuning}. This demonstrates that fine-tuning the \mbox{StyleGAN} backbone enhances the tri-plane features for decoding them into high-quality Gaussian Splatting scenes. 

\begin{figure}[htbp]
    \centering
    \includegraphics[width=1\linewidth]{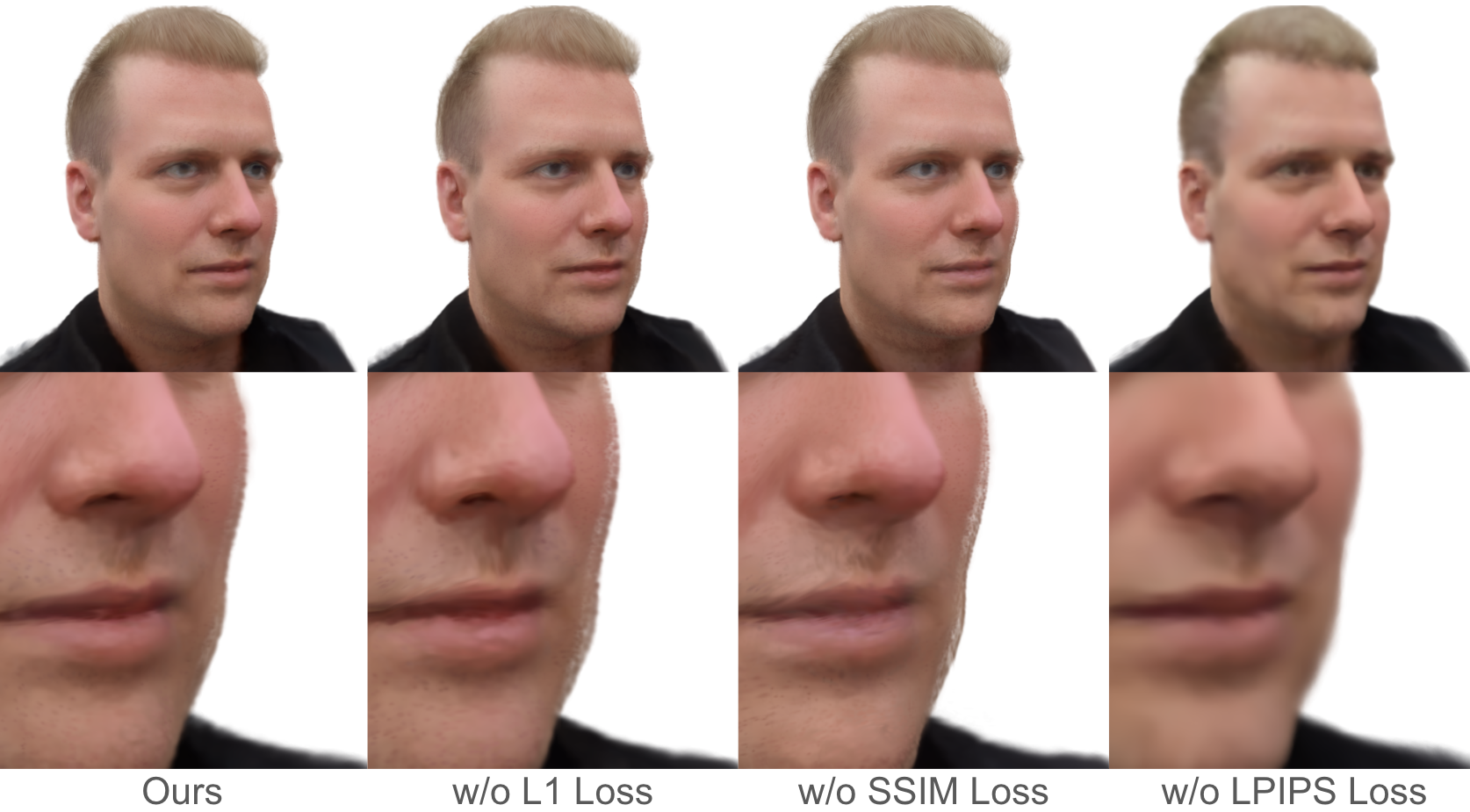}
    \caption{A visual comparison between decoders that have been trained while deactivating the respective loss function. }
    \label{fig:loss_ablation}
\end{figure}

\noindent{\textbf{Loss Functions:}} Training the decoder network requires appropriate loss functions that yield meaningful gradients. For our proposed decoder training, we employ a combination of several different loss functions. 
To better understand their individual impact, we conduct an ablation study by training multiple models, each with one loss function deactivated. The resulting renderings compared using the same ID can be seen in Figure \ref{fig:loss_ablation}. Here, the biggest difference is visible when deactivating the LPIPS loss. In this case, the rendering starts to become very blurry. This is surprising, given that L1 or SSIM are expected to penalize such blurry renderings. Instead, when disabling them, some artifacts are created at the edges. This hints that those loss functions help building the coarse geometry for the face, while LPIPS provides a gradient that creates fine structures.

\begin{table}[htbp]
    \centering
    \begin{tabular}{|l|ccc|}
        \hline
        Method & LPIPS $\downarrow$ & SSIM $\uparrow$ & ID Sim $\uparrow$ \\
        \hline
        Baseline          &	0.176	&	0.842	&	0.883 \\
        \hline

        w/o fine-tuning            &	0.188	&	0.837	&	0.858 \\
        \hline
        w/o L1 Loss & 0.175	&	0.841	&	0.881 \\
        w/o LPIPS Loss & 0.260	&	\textbf{0.859}	&	\textbf{0.885}\\
        w/o SSIM Loss &	\textbf{0.174}	&	0.832	&	0.880 \\
        w/o Sobel Loss &	0.175	&	0.839	&	0.880 \\
        w/o ID Loss &	0.176	&	0.842	&	0.827 \\
        \hline
    \end{tabular}
    \caption{Comparison of our baseline model with variants, each with a single component deactivated. While the baseline does not achieve the highest score for each metric, it offers a balanced trade-off along all three metrics combined.}
    \label{tab:finetuning}
\end{table}

%% file: sec/5_conclusion.tex
\section{Limitations \& Future Work}
\label{sec:limitations}

Since all target images we use to train our framework are generated by either PanoHead or EG3D, the output fidelity of our method is bounded by the fidelity of these 3D GANs. A possible approach to push the visual quality of our renderings closer to photorealism would be to train the entire pipeline, i.e.\ training the generator backbone alongside the decoder, from scratch in a GAN based end-to-end manner. This approach, while being straightforward in theory, is subject to some challenges including the localization of good initial positions for the Gaussians in 3D space and, especially, handling the unstable nature of adversarial training. We aim to tackle these challenges in succeeding works, using the general structure of this framework and the insights obtained while developing it as a foundation.

Moreover, we observe that the eyes of the faces in our scenes appear uncanny or blurry. As described in \ref{sec:qaulitative}, we expect to solve this issue by including view-dependent spherical harmonics in the future.



\section{Conclusion}
\label{sec:conclusion}

We have presented a framework that decodes tri-plane features of pre-trained 3D aware GANs for facial image generation like PanoHead or EG3D into scenes suitable for rendering with 3D Gaussian Splatting. This not only allows for rendering at up to 5 times higher frame rates with flexible image resolutions but also enables to export the resulting scenes into 3D software environments, allowing for realistic 3D asset creation in real-time. As our decoders show very high visual similarity to the 3D-aware target GANs, we are able to maintain high visual quality along interpolation paths, paving the way for applying GAN editing or GAN inversion methods to explicit 3D Gaussian Splatting scenes for the first time.
In an in-depth ablation study, we discuss all components of our method, providing a basis for future works dealing with decoding Gaussian Splatting attributes. For succeeding work, we plan to broaden our training scheme to be able to train a GAN in an adversarial training for the generation of scenes compatible with 3DGS, as mentioned in the previous section.



\section{Acknowledgements}
\label{sec:acknowledgements}
This research has partly been funded by the German Research Foundation (3DIL, grant no.\ 502864329), the European Union’s Horizon Europe research and innovation programme (Luminous, grant no.\ 101135724), and the German Ministry of Education and Research (MoDL, grant no.\ 01IS20044).

%% file: sec/X_suppl.tex
\clearpage
\setcounter{page}{1}
\maketitlesupplementary

\begin{strip}
    \centering
    \includegraphics[width=1.0\textwidth]{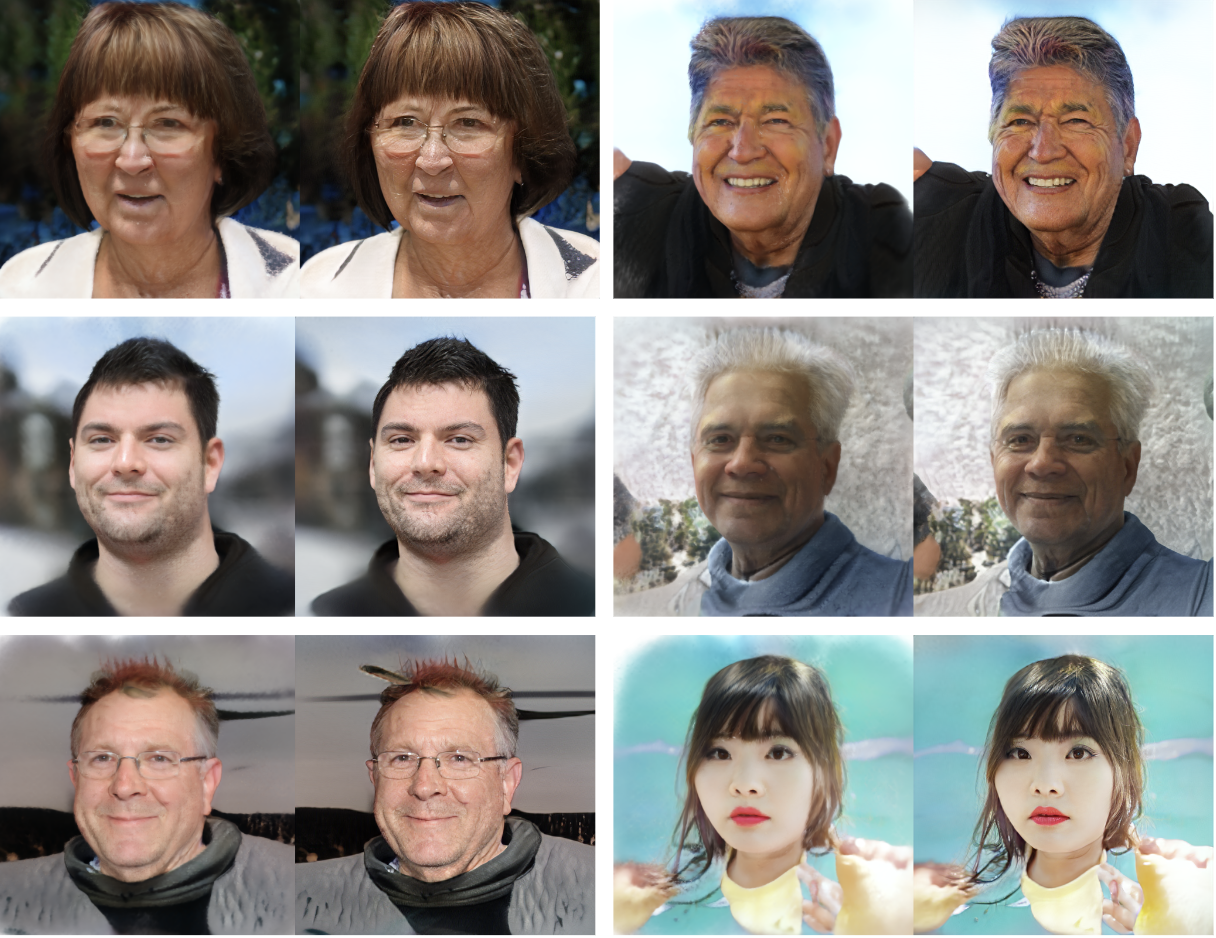}
    \captionof{figure}{Rendering examples for the EG3D-FFHQ decoder. Left shows the decoded Gaussian Splatting scene and right shows the original GAN rendering.
    }
    \label{fig:examples_ffhq}
\end{strip}

\begin{figure*}
    \centering
    \includegraphics[width=1\linewidth]{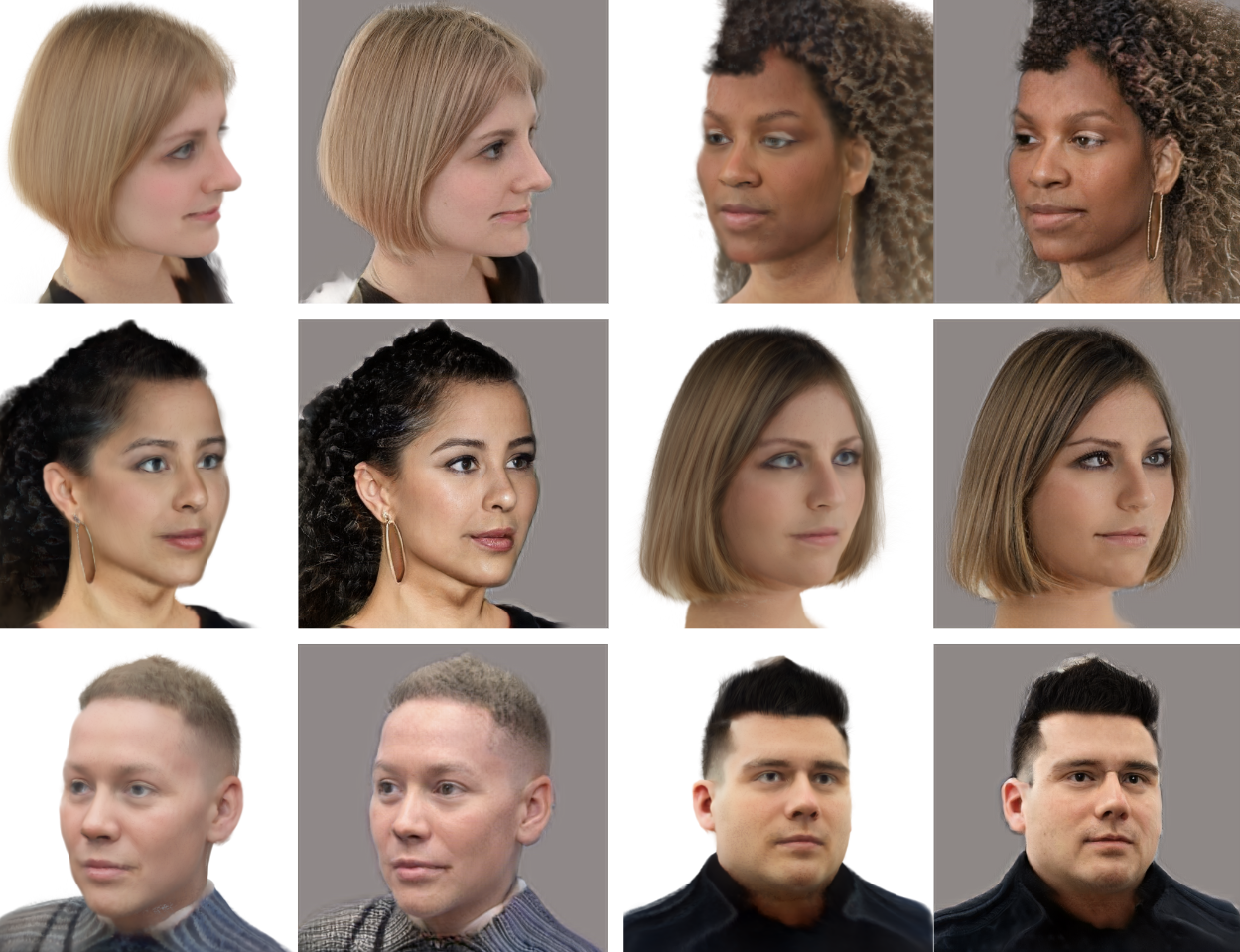}
    \caption{Rendering examples for the PanoHead decoder. Left shows the decoded Gaussian Splatting scene and right shows the original GAN rendering.}
    \label{fig:examples_panohead}
\end{figure*}

\begin{figure*}
    \centering
    \includegraphics[width=1\linewidth]{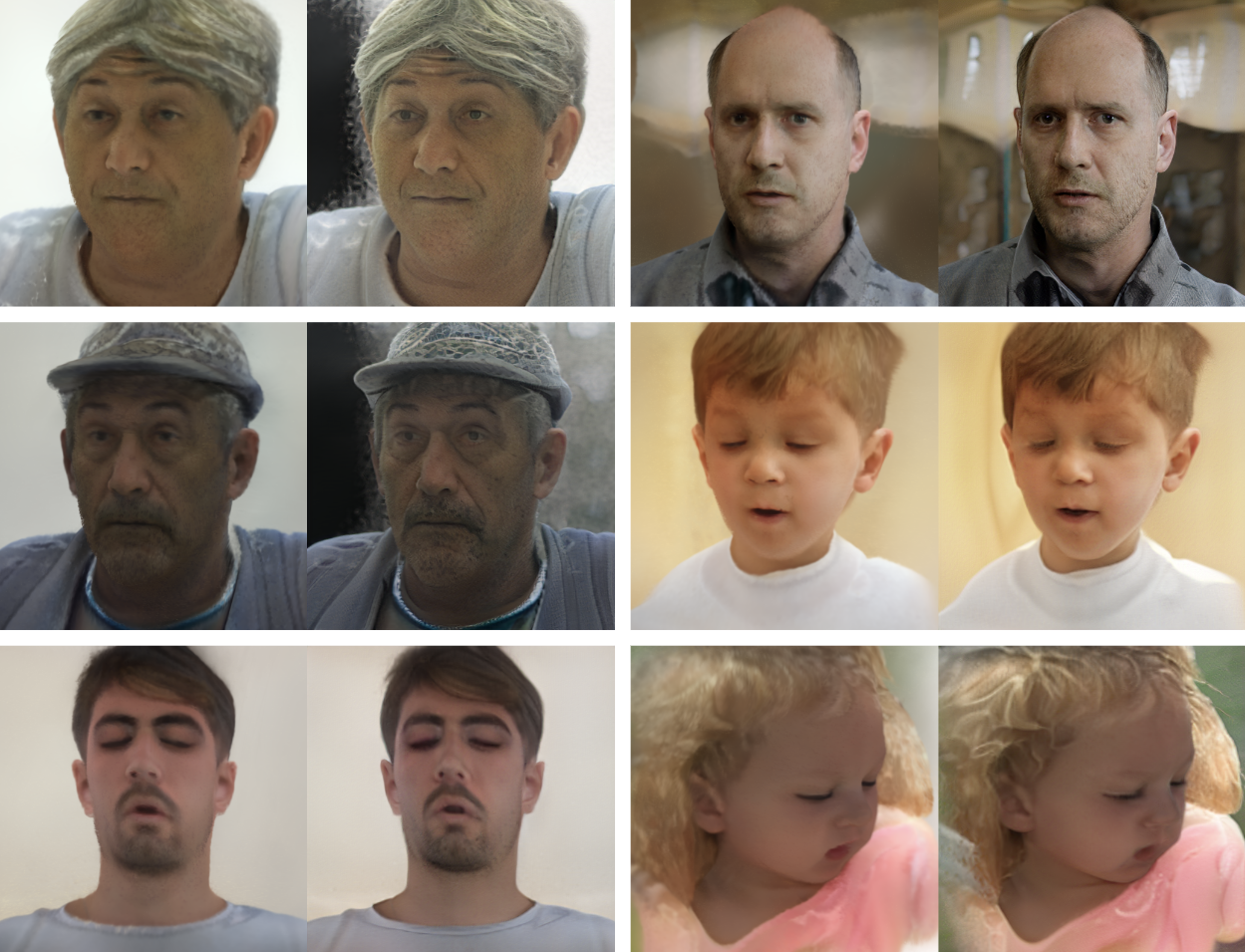}
    \caption{Rendering examples for the EG3D-LPFF decoder. Left shows the decoded Gaussian Splatting scene and right shows the original GAN rendering.}
    \label{fig:examples_lpff}
\end{figure*}

\begin{figure*}
    \centering
    \includegraphics[width=1\linewidth]{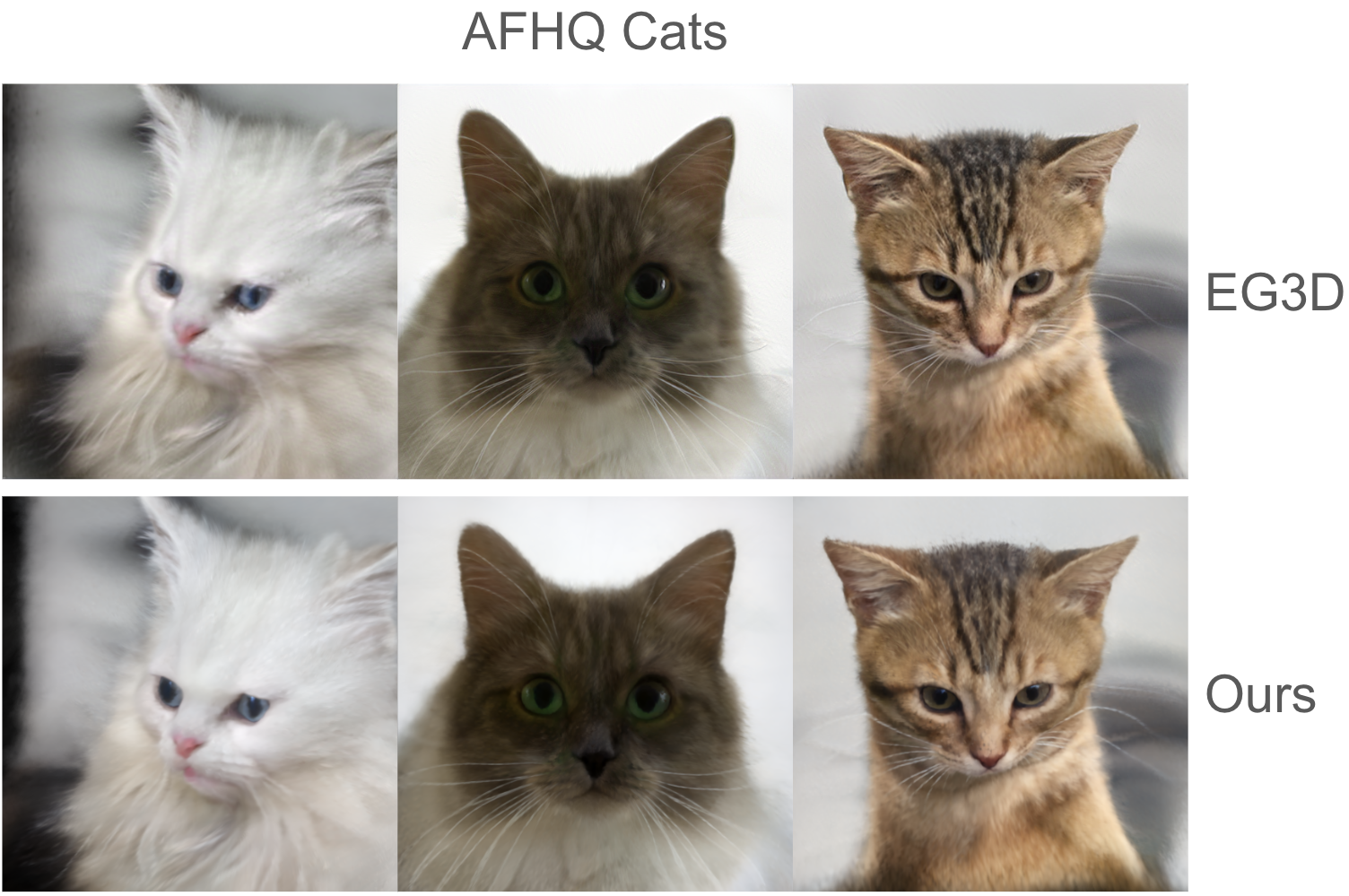}
    \caption{We also demonstrate that our method is not exclusively working for human heads. After disabling the ID similarity loss, we are able to decode a EG3D that was trained on AFHQ Cats.}
    \label{fig:examples_cats}
\end{figure*}